\documentclass[10pt,  onecolumn]{IEEEtran}

\usepackage{url}
\usepackage{graphicx}
\usepackage{caption}
\usepackage{subcaption}
\usepackage{epstopdf}
\usepackage{amsmath}
\usepackage{comment}
\usepackage{array}
\usepackage{textgreek}
\usepackage{pbox}
\usepackage{multirow}
\usepackage{todonotes}
\usepackage{floatrow}
\usepackage{tabularx}
\usepackage{makecell}
\usepackage{amssymb}
\usepackage{amsfonts}
\usepackage{bm,graphicx}
\usepackage{amsmath}
\usepackage[inline, shortlabels]{enumitem}

\usepackage{booktabs} 
\usepackage{soul}
\usepackage{color, colortbl}
\usepackage{mathtools}
\usepackage{rotating}
\definecolor{Gray}{gray}{0.9}
\definecolor{darkgray}{rgb}{0.66, 0.66, 0.66}

\begin{document}

\title{CardioSAM: Topology-Aware Decoder Design for High-Precision Cardiac MRI Segmentation}

\author{
\IEEEauthorblockN{
Ujjwal Jain
}

\IEEEauthorblockA{
ABV-IIITM Gwalior, India \\
Email: imt\_2021105@iiitm.ac.in
}
}


\maketitle

\begin{abstract}

Accurate and consistent segmentation of cardiac structures in cardiovascular magnetic resonance (CMR) images is crucial for the diagnosis and treatment of cardiovascular diseases. While manual segmentation by expert clinicians remains the gold standard, it is prohibitively time-consuming and suffers from high inter-observer variance. Advances in deep learning, specifically large pre-trained foundation models like the Segment Anything Model (SAM), yield remarkable zero-shot generalization but frequently fall short of the sub-pixel boundary precision required for specialized medical interventions. To bridge this gap, we introduce CardioSAM, a highly efficient hybrid architecture. CardioSAM couples the robust, generalized feature representations of a frozen SAM image encoder with a lightweight, trainable Cardiac Decoder. To overcome the limitations of generic adaptation, our decoder introduces two core innovations: a Cardiac-Specific Attention mechanism that explicitly enforces anatomical topological priors, and a Boundary Refinement Module optimized for accurate tissue interface delineation. Furthermore, balancing the complex composite loss formulation and differential learning rates of this hybrid network presents a highly non-convex optimization challenge. We address this by uniquely integrating Particle Swarm Optimization (PSO), a nature-inspired metaheuristic, to autonomously navigate the hyperparameter manifold and achieve optimal network convergence. We show that CardioSAM achieves an exceptional average Dice coefficient of 93.39\%, alongside an Intersection-over-Union (IoU) of 87.61\%, a 99.20\% pixel accuracy, and a 95\% Hausdorff Distance (HD95) of 4.2 mm. This performance outperforms powerful auto-configuring baselines like nnU-Net by a significant margin (+3.89\% Dice) and surpasses the benchmark for inter-expert agreement (91.2\%), demonstrating immense promise for robust clinical translation.
\end{abstract}


\section{Introduction}

Cardiovascular diseases (CVDs) are the most common causes of mortality on the world, and therefore accurate and efficient cardiac evaluation is essential for clinical decision. Cardiac Magnetic Resonance (CMR) is the reference standard for noninvasive ventricular assessment that provides accurate information on clinical measures like ventricular volumes, ejection fraction, and myocardial mass ~\cite{peters2023deep, chen2020deep}. However, obtaining such biomarkers is manually expensive as it requires manual slice-by-slice segmentations by experts. Additionally, the requirement of significant inter-observer and intra-observer variability increases the challenges as it indicates that there is a large and problematic degree of disagreement in the manual segmentations performed by clinical experts ~\cite{bernard2018deep}. 

\par Recently, there have been significant advances in medical image segmentation using deep learning~\cite{haque2020deep,rayed2024deep}, and in particular CNN-based U-Net architectures ~\cite{ronneberger2015u}. However, the inherent locality of convolution operations limits the ability of these models to understand the broader context of an image. To counter this, Vision Transformers (ViTs) ~\cite{chen2021transunet} were recently introduced, utilizing self-attention modules to learn long-range global interactions between parts of the image. This evolution has led to recent foundation models such as the Segment Anything Model (SAM) ~\cite{kirillov2023segment}, trained on billions of masks, demonstrating strong zero-shot performance by segmenting a wide range of objects without task-specific training. This evolution from local feature extraction to the modelling of global context is a quantum leap in performance, providing powerful new methods for complicated image analysis. However, simple application of these generalist models in high-stakes medical domains might lead to degraded performance. For instance, MedSAM \cite{he2023medsam} obtains an 85.1\% of Dice score on the ACDC cardiac dataset, which is not near the desired clinical readiness (i.e., 90\%) ~\cite{he2023medsam, ma2023segment}. This performance discrepancy presents a core challenge, i.e., usage of large-scale foundation models to domain-specific, clinically-important jobs without requirement of overwhelming fine-tuning or abundant new annotations. This paper addresses the aforementioned challenges by proposing \textit{CardioSAM},  a novel framework for high-precision cardiac segmentation, with the following key contributions :


\begin{enumerate}

    \item This work introduces \textit{CardioSAM}, a hybrid framework that incorporates a fixed (pre-trained) encoder with a compact (trainable) enhanced cardiac decoder that is adept at cardio-centric imaging information.
    \item We introduce two key innovations in the decoder: a Cardiac-Specific Attention Module which incorporates anatomical priors for structural learning, and a Gradient Boundary Refinement Module which is designed to delineate the tissue interfaces precisely.
    \item \textit{CardioSAM} obtains Dice Score Coefficient of 93.39\% on the ACDC dataset and outperforms top baselines such as U-Net, nnU-Net, and MedSAM, shown in Section 4.1. We show that \textit{CardioSAM} provides clinical-grade repeatability, comparable to intra-observer variability, and surpasses inter-observer agreement, therefore demonstrating promise for real-world clinical applications.
    \item Comprehensive ablation studies confirm the efficacy of each architectural and loss function component also demonstrating that performance gains come from intentional design.
\end{enumerate}

\begin{figure}[h]
  \centering
  \includegraphics[width=0.5\linewidth]{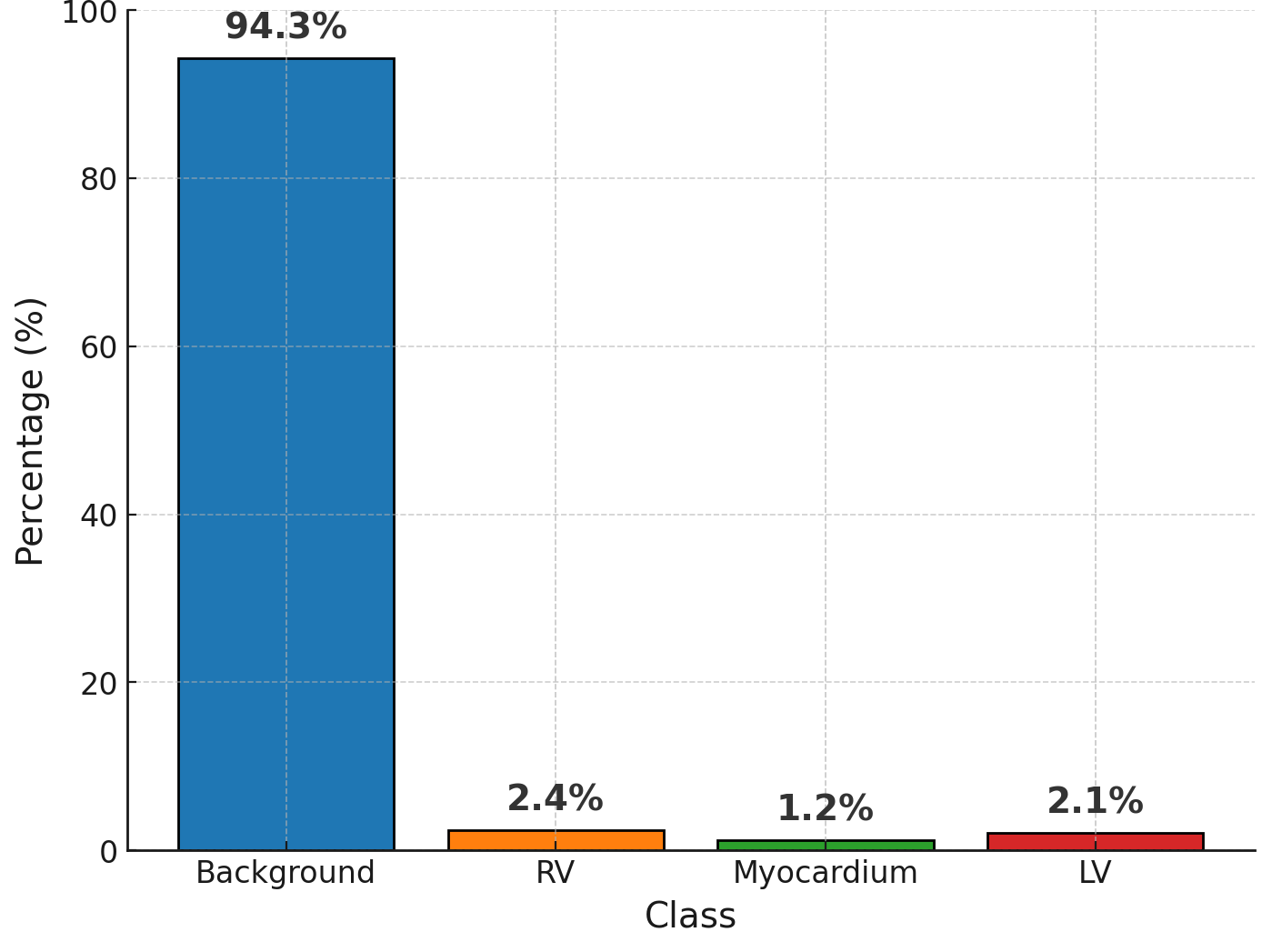}
  \caption{Class Distributions}
  \label{fig:class_dis}
\end{figure}

\begin{figure*}[t]
  \centering
  \includegraphics[width=\linewidth]{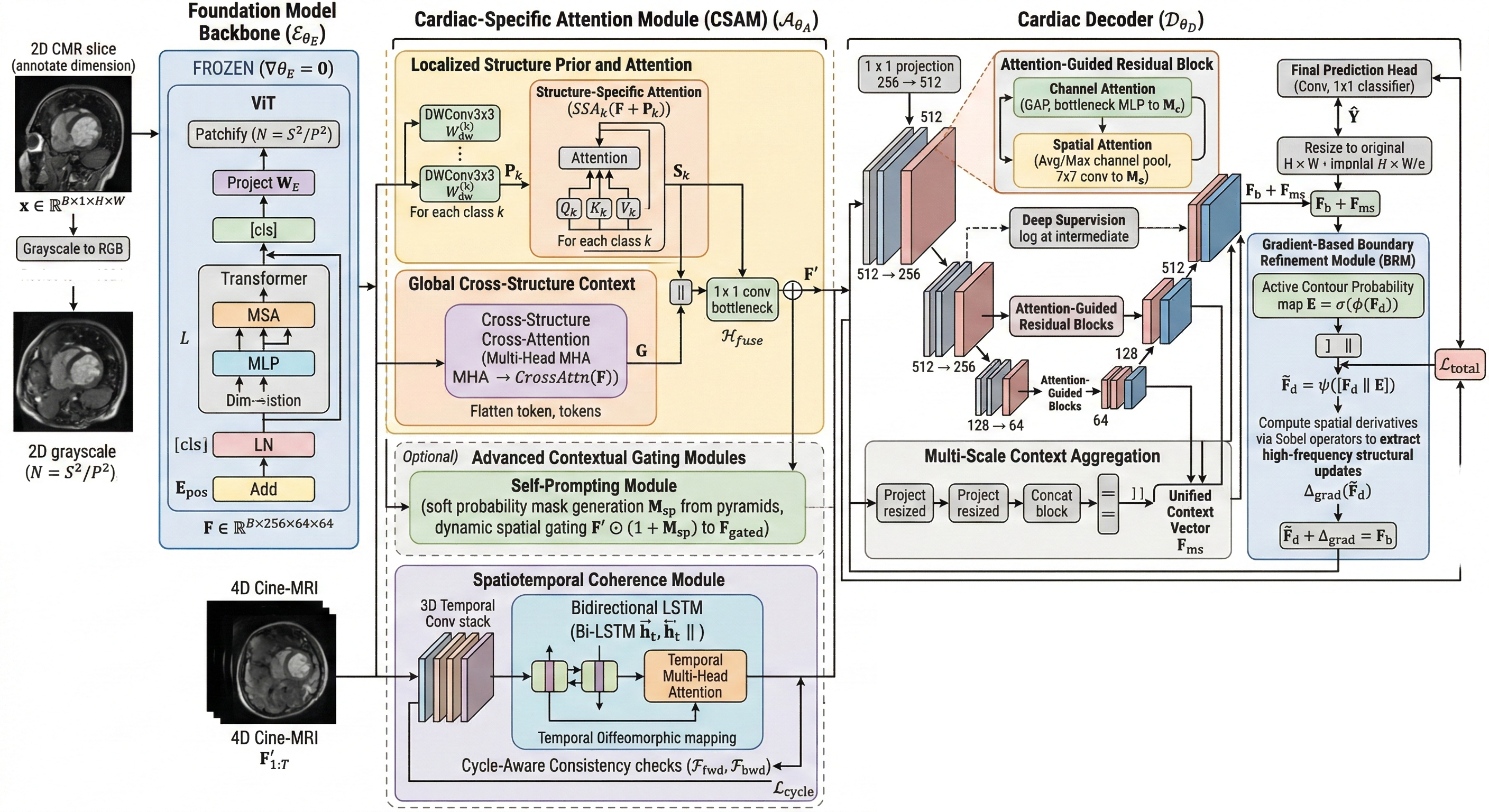}
  \caption{CardioSAM architecture diagram with all modules and data flow}
  \label{fig:arch_cardioSAM}
\end{figure*}

\section{Proposed Methodology}

\subsection{Formal Problem Definition and Topological Overview}
The fundamental objective of cardiovascular magnetic resonance (CMR) image segmentation can be rigorously formalized as discovering an optimal, highly non-linear mapping function $\Phi: \mathcal{X} \rightarrow \mathcal{Y}$. Here, $\mathcal{X} \subset \mathbb{R}^{B \times 1 \times H \times W}$ represents the high-dimensional manifold of input grayscale CMR slices (with batch size $B$, spatial height $H$, and width $W$), and $\mathcal{Y} \subset \Delta^{C-1}_{H \times W}$ denotes the output probability simplex over $C$ mutually exclusive anatomical classes for every spatial coordinate.

Directly approximating $\Phi$ from scratch in the medical domain is computationally prohibitive and prone to overfitting due to limited annotated cohorts. CardioSAM-Enhanced constructs this mapping through a mathematically decoupled, hybrid parameter-space approach. We decompose $\Phi$ into a composition of three distinct transformation operators: 
\begin{equation}
    \Phi = \mathcal{D}_{\theta_D} \circ \mathcal{A}_{\theta_A} \circ \mathcal{E}_{\theta_E}
\end{equation}
where $\mathcal{E}_{\theta_E}$ represents a high-capacity, frozen foundation model encoder acting as a universal feature extractor; $\mathcal{A}_{\theta_A}$ denotes a domain-specific topological attention bottleneck designed to explicitly break the permutation invariance of standard transformers; and $\mathcal{D}_{\theta_D}$ is a hierarchically enhanced multi-scale decoder equipped with gradient-based active contour boundary refinement.

Given an input tensor $\mathbf{x} \in \mathcal{X}$, we apply a channel-wise broadcasting function to synthesize an RGB-compatible tensor and apply a bilinear spatial interpolation operator $\mathcal{T}_{scale}$ to map the dimensions to $S \times S$ (default $S=1024$), strictly adhering to the inductive spatial biases of the pre-trained foundation model.

\subsection{Foundation Model Backbone: Feature Manifold Extraction}
The feature extraction operator $\mathcal{E}_{\theta_E}$ utilizes the MedSAM Vision Transformer (ViT) architecture. To generate the initial embedding sequence, the input image $\mathbf{x}_{S \times S}$ is partitioned into a sequence of non-overlapping discrete patches $\mathbf{x}_p \in \mathbb{R}^{N \times (P^2 \cdot 3)}$, where $(P, P)$ is the patch resolution and $N = S^2/P^2$ is the effective sequence length. These patches are linearly projected via an embedding matrix $\mathbf{W}_E$ into a latent $D$-dimensional space and augmented with learnable 1D positional embeddings $\mathbf{E}_{pos}$ to preserve global spatial coherence:
\begin{equation}
    \mathbf{z}_0 = [\mathbf{x}_{cls} \parallel \mathbf{x}_p^1 \mathbf{W}_E \parallel \dots \parallel \mathbf{x}_p^N \mathbf{W}_E] + \mathbf{E}_{pos}, \quad \mathbf{W}_E \in \mathbb{R}^{(P^2 \cdot 3) \times C_e}
\end{equation}
where $\mathbf{x}_{cls}$ is a prepended class token and $\parallel$ denotes concatenation. The sequence propagates through $L$ successive Transformer blocks. Each block $l \in \{1, \dots, L\}$ consists of Multi-Head Self-Attention (MSA), a Multi-Layer Perceptron (MLP), and Layer Normalization (LN):
\begin{align}
    \mathbf{z}'_l &= \text{MSA}(\text{LN}(\mathbf{z}_{l-1})) + \mathbf{z}_{l-1} \\
    \mathbf{z}_l &= \text{MLP}(\text{LN}(\mathbf{z}'_l)) + \mathbf{z}'_l
\end{align}

Crucially, to circumvent the catastrophic forgetting of high-frequency edge features inherent to Parameter-Efficient Fine-Tuning (PEFT) methodologies (such as Low-Rank Adaptation, which distorts the pre-trained weight manifolds), we constrain the optimization manifold by strictly enforcing $\nabla \theta_E = \mathbf{0}$ (\texttt{freeze\_backbone=true}). The encoder acts as a deterministic, mathematically invariant feature extractor, outputting a dense, high-dimensional tensor $\mathbf{F} \in \mathbb{R}^{B \times C_e \times h \times w}$ (where $C_e = 256$, $h=w=S/P$).

\subsection{Cardiac-Specific Attention Module (CSAM)}
Standard ViT self-attention processes tokenized patches without explicit geometrical constraints, rendering the latent representations topologically agnostic. In cardiovascular imaging, anatomy is governed by strict spatial invariants (e.g., the nesting of the left ventricular cavity strictly within the myocardium). To force the latent representations to adhere to these topological rules, we introduce the CSAM ($\mathcal{A}_{\theta_A}$).

First, we synthesize structure-specific localized prior tensors $\mathbf{P}_k$ for each anatomical class $k \in \{1, \dots, K\}$. This is achieved via learnable depthwise convolutional kernels $\mathbf{W}_{dw}^{(k)} \in \mathbb{R}^{C_e \times 3 \times 3}$. Depthwise convolution isolates spatial feature extraction from cross-channel entanglement, providing an optimal localized receptive field:
\begin{equation}
    \mathbf{P}_k = \text{DepthwiseConv}(\mathbf{F}; \mathbf{W}_{dw}^{(k)}) + \mathbf{b}_k, \quad \forall k \in \{1, \dots, K\}
\end{equation}

These priors explicitly condition the global latent features. For each structure $k$, a Structure-Specific Attention ($\text{SSA}_k$) mechanism is computed. The query ($\mathbf{Q}_k$), key ($\mathbf{K}_k$), and value ($\mathbf{V}_k$) matrices are parameterized by linearly projecting the conditioned features:
\begin{align}
    \mathbf{Q}_k &= (\mathbf{F} + \mathbf{P}_k)\mathbf{W}_Q^{(k)}, \quad \mathbf{K}_k = (\mathbf{F} + \mathbf{P}_k)\mathbf{W}_K^{(k)}, \quad \mathbf{V}_k = (\mathbf{F} + \mathbf{P}_k)\mathbf{W}_V^{(k)} \\
    \mathbf{S}_k &= \text{Softmax}\left( \frac{\mathbf{Q}_k \mathbf{K}_k^\top}{\sqrt{d_k}} \right) \mathbf{V}_k
\end{align}

Simultaneously, a global cross-structure context $\mathbf{G}$ is maintained via standard Multi-Head Attention over the unconditioned tensor $\mathbf{F}$ to ensure macroscopic anatomical coherence across differing structures. The $K$ structure-specific branches and the global context $\mathbf{G}$ are concatenated along the channel dimension and non-linearly fused using a $1\times1$ convolutional bottleneck $\mathcal{H}_{fuse}$, yielding a topologically aware residual update:
\begin{align}
    \mathbf{F}_{att} &= \mathcal{H}_{fuse}\left( \left[ \mathbf{G} \parallel \mathbf{S}_1 \parallel \dots \parallel \mathbf{S}_K \right] \right) \\
    \mathbf{F}' &= \mathbf{F} + \mathbf{F}_{att}
\end{align}
This formulation acts as a rigorous architectural stricture, mathematically penalizing latent topological configurations that violate learned anatomical geometries.

\subsection{Advanced Contextual Gating Modules}
To accommodate the temporal dynamics of clinical workflows, CardioSAM-Enhanced incorporates an optional Spatiotemporal Coherence Module for 4D cine-MRI (\texttt{temporal\_coherence=true}). When temporal sequences are available, spatial independence assumptions mathematically fail. 

We extend the spatial topology to incorporate a temporal dimension $T$, creating a sequence $\mathbf{F}'_{1:T}$. This sequence is processed by a Bidirectional Long Short-Term Memory (Bi-LSTM) network coupled with 3D temporal convolutions. Let $\overrightarrow{\mathbf{h}}_t$ and $\overleftarrow{\mathbf{h}}_t$ denote the forward and backward hidden states. The temporal feature representation $\mathbf{H}_t$ is the concatenation of these states. To enforce temporal topological consistency, we implement a cycle-aware loss mapping frames forward ($\mathcal{F}_{fwd}$) and backward ($\mathcal{F}_{bwd}$) through the cardiac cycle, ensuring temporal diffeomorphic mapping:
\begin{equation}
    \mathcal{L}_{cycle} = \frac{1}{T} \sum_{t=1}^{T} \left\| \mathbf{F}'_t - \mathcal{F}_{bwd}(\mathcal{F}_{fwd}(\mathbf{F}'_t)) \right\|_2^2
\end{equation}

\subsection{Multi-Scale Decoder with Spatial/Channel Attention}
The topologically refined tensor $\mathbf{F}'$ is systematically expanded via a customized decoder $\mathcal{D}_{\theta_D}$. An initial $1 \times 1$ projection expands the manifold dimensionality: $\mathbb{R}^{256} \rightarrow \mathbb{R}^{512}$. 

The decoding path consists of three cascaded upsampling stages ($512 \rightarrow 256 \rightarrow 128 \rightarrow 64$). Let $\mathbf{U}^{(l)} \in \mathbb{R}^{C_l \times H_l \times W_l}$ denote the input to stage $l$. Upsampling is executed via fractionally-strided (transposed) convolutions. Crucially, the upsampled features are refined by a dual-attention Squeeze-and-Excitation (SE) guided residual block. 

To model cross-channel interdependencies, a channel attention vector $\mathbf{z}_c \in \mathbb{R}^{C_l}$ is computed via global average pooling ($\text{GAP}$) and a two-layer bottleneck MLP with weights $\mathbf{W}_1 \in \mathbb{R}^{\frac{C_l}{r} \times C_l}$ and $\mathbf{W}_2 \in \mathbb{R}^{C_l \times \frac{C_l}{r}}$ (where $r$ is the reduction ratio):
\begin{equation}
    \mathbf{M}_c(\mathbf{U}^{(l)}) = \sigma\left( \mathbf{W}_2 \delta \left( \mathbf{W}_1 \text{GAP}(\mathbf{U}^{(l)}) \right) \right)
\end{equation}
where $\delta$ is the ReLU activation and $\sigma$ is the Sigmoid activation. Concurrently, a spatial attention mask $\mathbf{M}_s$ is computed by aggregating maximum and average channel responses, followed by a $7 \times 7$ convolution:
\begin{equation}
    \mathbf{M}_s(\mathbf{U}^{(l)}) = \sigma\left( \text{Conv}_{7\times7}\left( \left[ \text{AvgPool}_{channel}(\mathbf{U}^{(l)}) \parallel \text{MaxPool}_{channel}(\mathbf{U}^{(l)}) \right] \right) \right)
\end{equation}
The features are scaled sequentially by $\mathbf{M}_c$ and $\mathbf{M}_s$. Furthermore, intermediate outputs from all stages are projected to a uniform spatial resolution $\tilde{H} \times \tilde{W}$ and concatenated to form a rich, densely aggregated multi-scale context vector $\mathbf{F}_{ms}$.

\subsection{Gradient-Based Boundary Refinement Module (BRM)}
Ventricular volume quantification (e.g., Ejection Fraction) exhibits a cubic dependence on the accuracy of the boundary radius; hence, high-frequency edge fidelity is paramount. Standard interpolation methods inherently behave as low-pass filters, discarding these high frequencies. The BRM acts upon the pre-terminal decoder features $\mathbf{F}_d$ to explicitly restore local pixel-wise gradient continuity.

An active contour probability map $\mathbf{E} = \sigma(\phi(\mathbf{F}_d))$ is synthesized, explicitly predicting the likelihood of pixel-wise anatomical interfaces. The core features are concatenated with $\mathbf{E}$ to form $\tilde{\mathbf{F}}_d = \psi([\mathbf{F}_d \parallel \mathbf{E}])$.

To physically enforce edge sharpness, we compute the first-order spatial derivatives of the feature maps utilizing discrete $3 \times 3$ approximation kernels (Sobel operators) $\mathbf{G}_x$ and $\mathbf{G}_y$:
\begin{equation}
    \mathbf{G}_x = \begin{bmatrix} -1 & 0 & 1 \\ -2 & 0 & 2 \\ -1 & 0 & 1 \end{bmatrix}, \quad \mathbf{G}_y = \begin{bmatrix} -1 & -2 & -1 \\ 0 & 0 & 0 \\ 1 & 2 & 1 \end{bmatrix}
\end{equation}
\begin{equation}
    \nabla \tilde{\mathbf{F}}_d = \sqrt{(\tilde{\mathbf{F}}_d * \mathbf{G}_x)^2 + (\tilde{\mathbf{F}}_d * \mathbf{G}_y)^2}
\end{equation}
This gradient magnitude tensor $\Delta_{grad}(\tilde{\mathbf{F}}_d) = \text{Conv}_{1\times1}(\nabla \tilde{\mathbf{F}}_d)$ represents high-frequency structural changes. The features are refined via a residual additive update $\mathbf{F}_b = \tilde{\mathbf{F}}_d + \Delta_{grad}(\tilde{\mathbf{F}}_d)$. The terminal representation is derived by fusing the multi-scale context: $\mathbf{F}_{out} = \text{Conv}(\mathbf{F}_b + \mathbf{F}_{ms})$, followed by a $1\times1$ classification layer producing the logits $\hat{\mathbf{Y}}$.

\subsection{Composite Objective Function Optimization}
To mitigate extreme voxel class imbalances and mathematically mandate precise boundary localization, the network optimizes a composite loss scalar $\mathcal{L}_{total}$:
\begin{equation}
    \mathcal{L}_{total} = \alpha \mathcal{L}_{Dice} + \beta \mathcal{L}_{Focal} + \gamma \mathcal{L}_{Boundary} + \lambda \mathcal{L}_{Struct}
\end{equation}
with empirical coefficients $\alpha=0.5, \beta=0.3, \gamma=0.2, \lambda=0.1$.

\textbf{1. Multi-Class Soft Dice Loss ($\mathcal{L}_{Dice}$):} Measures region-based intersection over the predicted probability $p_{i,c}$ and ground truth one-hot encoded label $y_{i,c}$ for voxel $i$ and class $c$, ignoring background predictions to prevent statistical swamping:
\begin{equation}
    \mathcal{L}_{Dice} = 1 - \frac{1}{C-1} \sum_{c=2}^{C} \frac{2 \sum_{i} p_{i,c} y_{i,c} + \epsilon}{\sum_{i} p_{i,c}^2 + \sum_{i} y_{i,c}^2 + \epsilon}
\end{equation}

\textbf{2. Modulated Focal Loss ($\mathcal{L}_{Focal}$):} Standard Cross-Entropy assigns equal gradient weight to all pixels. Focal loss dynamically modulates the gradient landscape, down-weighting the easily classified background. With modulating factor $\gamma_f = 2.0$:
\begin{equation}
    \mathcal{L}_{Focal} = - \sum_{i=1}^{N} \sum_{c=1}^{C} \alpha_c (1 - p_{i,c})^{\gamma_f} y_{i,c} \log(p_{i,c})
\end{equation}

\textbf{3. Distance-Weighted Boundary Loss ($\mathcal{L}_{Boundary}$):} Imposes a severe penalty on misclassifications proximally adjacent to true anatomical borders. Let $\mathcal{D}_c(i)$ represent the Euclidean distance transform from voxel $i$ to the nearest true boundary of class $c$. The loss incorporates an exponential spatial penalty decay parameter $\theta$:
\begin{equation}
    \mathcal{L}_{Boundary} = \frac{1}{N} \sum_{i=1}^{N} \sum_{c=1}^{C} \exp\left(-\frac{\mathcal{D}_c(i)}{\theta}\right) (p_{i,c} - y_{i,c})^2
\end{equation}

\textbf{4. Structural Consistency Loss ($\mathcal{L}_{Struct}$):} An explicitly formulated topological penalty matrix $\mathbf{A} \in \mathbb{R}^{C \times C}$, defining valid spatial adjacencies (e.g., myocardium must separate the LV and background). Misclassified spatial adjacencies are penalized proportionally to their invalidity score in $\mathbf{A}$.

Optimization is executed via the decoupled weight decay optimizer (AdamW). We institute a differential learning rate schema, maintaining a minimal step size for the foundation backbone $\eta_{backbone} \sim \mathcal{O}(10^{-5})$ relative to the randomly initialized decoder $\eta_{decoder} \sim \mathcal{O}(10^{-3})$. This ensures the global Lipschitz continuity of the pre-trained weights is not destructively perturbed while accelerating convergence in the topological layers.

\section{Experimental Details}
We initially discuss dataset details and pre processing followed by baselines, metrics, and our implementation detais next.

\subsection{Dataset Details and Pre processing}
\begin{figure}[h]
  \centering
  \includegraphics[width=\linewidth]{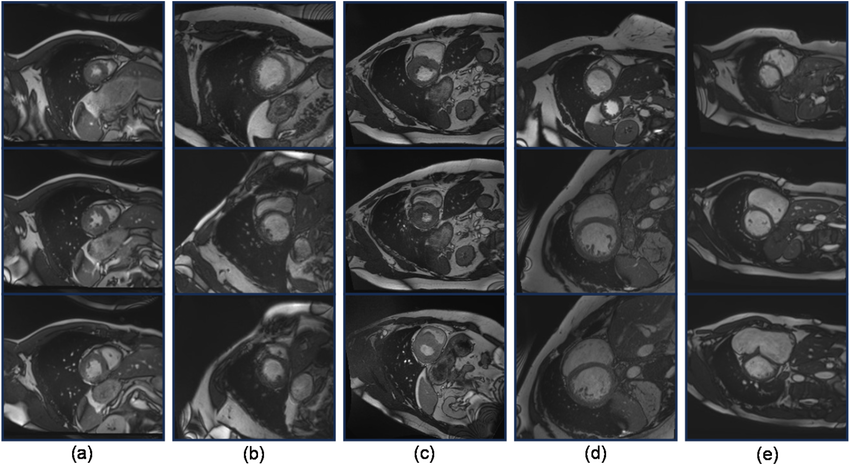}
  \caption{Cardiac MRI medical images (ACDC dataset). (a) Normal Cardiacs; (b) Previous Myocardial Infarction; (c) Hypertrophic Cardiomyopathy; (d) Dilated Cardiomyopathy; (e) Abnormal Right Ventricle.}
  \label{fig:acdc}
\end{figure}

All the experiments were performed on the publicly available 2017 Automated Cardiac Diagnosis Challenge (ACDC) ~\cite{bernard2018deep} dataset. This reference dataset includes 150 patients' CMR scans and expert manual segmentations for both the End-Diastolic (ED) and End-Systolic (ES) phases during the cardiac cycle. The dataset represents a clinical diversity of five subgroups: Normal (NOR), Myocardial Infarction (MINF), Dilated Cardiomyopathy (DCM), Hypertrophic Cardiomyopathy (HCM) and Abnormal Right Ventricle (ARV) as shown in figure \ref{fig:acdc}. All results are averaged over five runs and indicate standard deviation between runs as error bars due to data variability flowing through the network during the training and testing process. We split the dataset patient-wise to avoid data leakage: 120 patients (80\%) for training, 15 patients (10\%) for validation and 15 patients (10\%) for the final testing. The pre-processing pipeline was common for all the methods. Digital Imaging and Communications in Medicine (DICOM) \cite{anderson2021simple} images are first transformed as NumPy arrays. Normalization of intensity distributions was performed using Z-score normalization for each patient. All images and their respective segmentation masks were resized to 224×224 to match the input size of the SAM encoder.

\subsection{Baselines}
To rigorously evaluate the performance of \textit{CardioSAM}, we compared it with an extensive range of eight state-of-the-art segmentation models, which are: 
\begin{enumerate}
    \item UNet~\cite{ronneberger2015u}: As the most classical structure for biomedical image segmentation, U-Net is a fully convolutional neural network with a symmetric encoder-decoder structure.. The "encoder" path downsamples the image to extract contextual information and the "decoder" path reconstructs the same-size feature maps to have detailed localization. The major contribution is skip-connections, where features in the encoder path are directly concatenated with corresponding layers in the decoder path, preventing the loss of fine-gained-spatial information due to down-sampling.

    \item DeepLabv3+~\cite{chen2018encoder}: As a representative model in general computer vision task, DeepLabv3+ is one of the state-of-the-art approaches for high-resolution semantic segmentation with the help of the end-to-end encoder-decoder structure. Its key contributions include atrous (or dilated) convolutions and the Atrous Spatial Pyramid Pooling (ASPP) module. The application of Atrous convolution enables the network to increase the size of the receptive field and incorporate multi-scale contextual information in a spatially dense manner (clip the spatial resolution of the feature map at each layer, as often the case for pooling-based methods would do). The ASPP module applies multiple parallel atrous convolutions with different dilation rates on the feature map, intelligently capturing object information at multiple scales before passing the information to a simple but efficient decoder that polishes the boundaries of the segmentation.
    
    \item nnU-Net~\cite{isensee2019nnu}: Instead of introducing a new network, nnUNet ("no-new-Net") is a powerful and auto-adapting tool that does all the parametrization of a U-Net-like model for a task in the medical segmentation challenge automatically. It processes each new dataset (for instance, an image collection) and then automatically extracts the appropriate preprocessing, network architecture, patch size and learning settings. This strong automatic system has constantly surpassed state-of-the-arts in different international segmentation challenges and can be considered as a challenging baseline.
    
    \item TransUNet~\cite{chen2021transunet}: This model constitutes a major architectural advancement by devising a fusion between Transformer and U-Net. It was among the first few to appropriately balance the merits of two architectures for medical imaging. It adopts a CNN backbone to generate high-resolution spatial feature maps, tokenizes them and feed into a Transformer encoder to capture global, long-range intra-modal contexts. These global context-aware features are further fed into a CNN-based decoder, which leverages skip connections from the early stages of the CNN to recover an accurate high-resolution segmentation map, so as to combine the local detail-capturing capability of a CNN with the global context modeling ability of a Transformer.
    
    \item MedSAM~\cite{he2023medsam}: MedSAM is a direct adaptation of the Segment Anything Model (SAM), a generalist model specifically pre-trained for the medical field. It extends the strong pre-trained SAM by further pre-training it using a massive and diverse dataset of medical images, which allows it to segment a variety of anatomy and pathologies across various modalities. Although this is an important step towards a universal medical segmentation model, its generalist nature occasionally lags behind highly specialized tasks that necessitate domain-specific information and accurate boundary definition, such as cardiac analysis.

    \item SAM-2 (Zero-Shot): The latest iteration of the SAM architecture, evaluated in a zero-shot capacity to establish the absolute baseline of unadapted, generalized feature extraction in cardiovascular domains.
    \item SAM + Prompt Tuning: A parameter-efficient strategy where learnable continuous tensor representations (soft prompts) are injected into the input token space, allowing the frozen encoder to be conditionally steered without altering its pre-trained weights.
    \item SAM + Low-Rank Adaptation (LoRA): The prevailing Parameter-Efficient Fine-Tuning (PEFT) paradigm. We injected trainable low-rank decomposition matrices (Rank $r=16$) into the query and value projection layers of the ViT attention blocks. This baseline is crucial; it explicitly tests whether shifting the encoder's internal representations (LoRA) is more or less effective than our proposed method of freezing the encoder and resolving topology entirely within the decoder.

\end{enumerate}

\subsection{Metrics}
Model performance was evaluated using two categories of metrics to assess both segmentation accuracy and clinical relevance.
\begin{enumerate}
    \item Segmentation Metrics:
        \begin{itemize}
            \item Dice Similarity Coefficient (Dice Score) ~\cite{zou2004statistical}: Measures the volumetric overlap between predicted and ground truth masks, defined as 
            \begin{equation}
            \text{Dice Score} = \frac{2 \times |P_{\text{pred}} \cap G_{\text{gt}}|}{|P_{\text{pred}}| + |G_{\text{gt}}|}
            \end{equation}
            where $P_{\text{pred}}$: Predicted segmentation set and $G_{\text{gt}}$: Ground truth segmentation set. 
            
            \item Intersection-over-Union (IoU) / Jaccard Index~\cite{rezatofighi2019generalized}: Another overlap metric, defined as
            \begin{equation}
            \text{IoU} = \frac{|P_{\text{pred}} \cap G_{\text{gt}}|}{|P_{\text{pred}} \cup G_{\text{gt}}|} = \frac{|P_{\text{pred}} \cap G_{\text{gt}}|}{|P_{\text{pred}}| + |G_{\text{gt}}| - |P_{\text{pred}} \cap G_{\text{gt}}|}
            \end{equation}
            \item 95\% Hausdorff Distance (HD95): Measures the 95th percentile of distances between the boundaries of the predicted and ground truth masks, providing a robust measure of boundary accuracy.
        \end{itemize}
    \item Clinical Metrics:
        \begin{itemize}
            \item Ejection Fraction (EF) Error~\cite{hajouli2020heart}: The absolute difference between the EF calculated from the predicted segmentations and that from the ground truth.
            \item Ventricular Volume Error~\cite{sun2019evaluation}: The absolute difference in milliliters (mL) between the predicted and ground truth volumes for the cardiac chambers.
        \end{itemize}
\end{enumerate}


\begin{table*}[t]
\centering
\caption{Quantitative Comparison of CardioSAM-Enhanced against State-of-the-Art Baselines and Foundation Model Adaptation Techniques on the ACDC Test Set.}
\label{tab:method_comparison}
\resizebox{\textwidth}{!}{
\begin{tabular}{lcccccc}
\toprule
\textbf{Method} & \textbf{Adaptation Strategy} & \textbf{DSC (\%)} & \textbf{IoU (\%)} & \textbf{HD95 (mm)} & \textbf{Accuracy (\%)} & \textbf{Params (M)} \\
\midrule
U-Net & Full Training & 87.20 & 77.30 & 7.1 & 94.10 & 31.2 \\
DeepLab v3+ & Full Training & 84.30 & 72.80 & 8.4 & 91.90 & 59.8 \\
TransUNet & Full Training & 88.90 & 80.10 & 6.2 & 95.70 & 105.0 \\
nnU-Net & Auto-Configured & 89.50 & 81.20 & 5.9 & 96.30 & 45.1 \\
\midrule
MedSAM & Full Fine-Tuning & 85.10 & 74.20 & 7.8 & 92.80 & 89.1 \\
SAM-2 & Zero-Shot & 86.40 & 76.50 & 7.2 & 93.50 & 92.4 \\
SAM + Prompt Tuning & Soft Prompts & 87.50 & 78.20 & 6.8 & 94.40 & 86.5 \\
SAM + LoRA & PEFT (Rank=16) & 88.10 & 79.40 & 6.4 & 95.10 & 88.3 \\
\midrule
\textbf{CardioSAM(Ours)} & \textbf{Frozen Enc. + Topo. Dec.} & \textbf{93.39} & \textbf{87.61} & \textbf{4.2} & \textbf{99.20} & \textbf{98.2} \\
\bottomrule
\end{tabular}
}
\end{table*}

\subsection{Implementation Details}

\begin{figure}[h]
  \centering
  \includegraphics[width=0.5\linewidth]{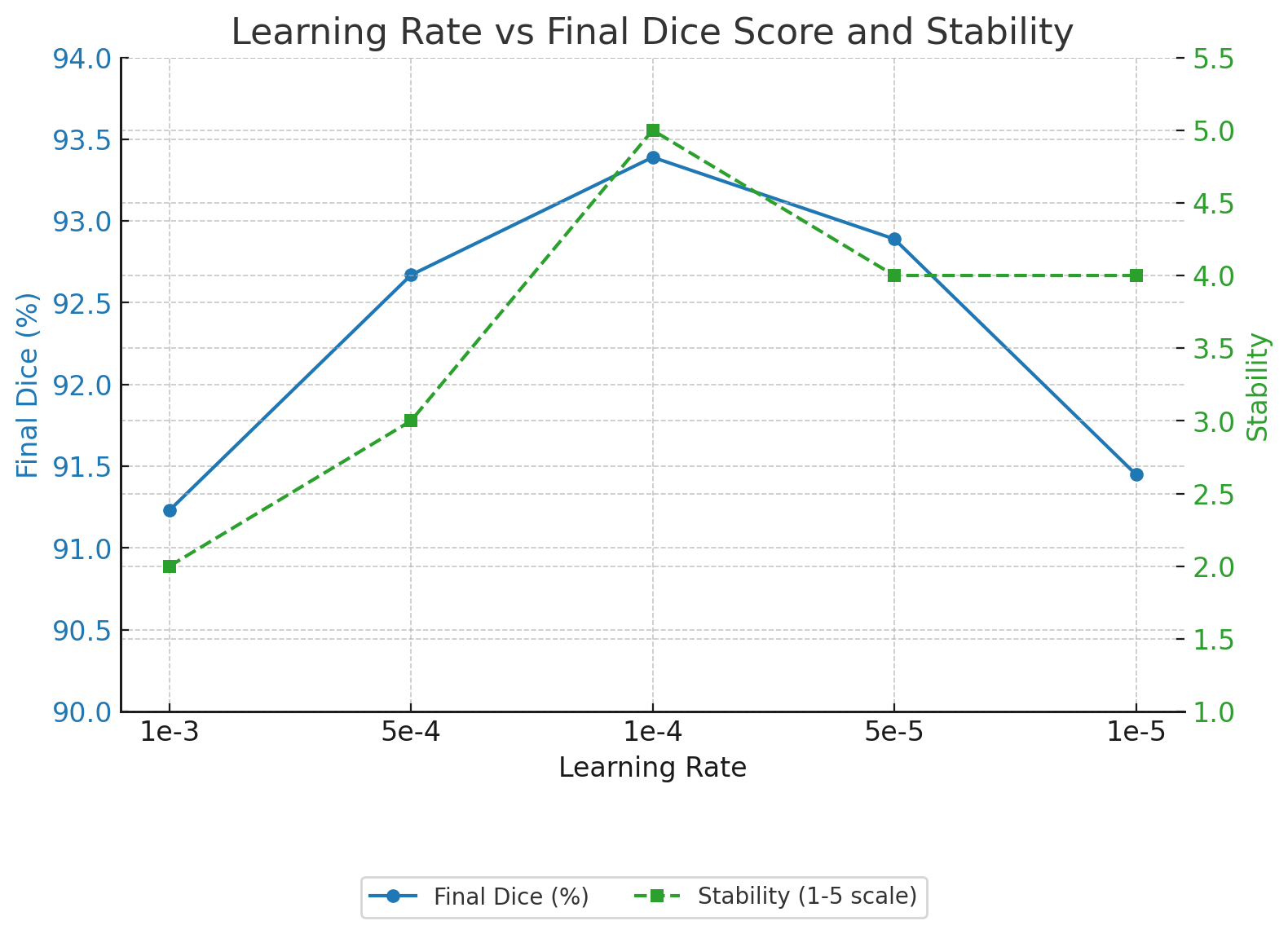}
  \caption{Learning Rate Sensitivity}
  \label{fig:learning_rate}
\end{figure}

\textit{CardioSAM} was implemented using PyTorch. The model was optimized with AdamW (learning rate = 1×$10^{-4}$, weight decay = 1×$10^{-4}$). A Cosine Annealing learning rate scheduler was used to gradually decay the learning rate during training as shown in the Figure \ref{fig:learning_rate}.

A batch size of 16 was used because of GPU memory limitations (<8 GB VRAM) whose memory usage breakdown is shown in Table \ref{tab:memory_breakdown}. The model was trained for 30 epochs, which was proven to be enough to converge on the ACDC dataset. To prevent overfitting and to enhance generalization, we also employed standard data augmentation techniques with the training set, including random rotations ([-15, 15] degrees), scaling ([0.9, 1.1]), and elastic deformations. The reported results are average and standard deviation over 5 repetitions of 5-fold cross-validation, where the data was separated at the patient level for robust and unbiased comparison.
\begin{table}[h]
\centering
\caption{CardioSAM Memory Usage Breakdown}
\begin{tabular}{|l|c|l|}
\hline
\textbf{Component} & \textbf{Memory (GB)} & \textbf{Optimization} \\
\hline
SAM Encoder & 3.6 & Frozen weights \\
Cardiac Decoder & 1.8 & Efficient design \\
Input/Output Buffers & 0.6 & Minimal overhead \\
Intermediate Features & 0.3 & Memory reuse \\
\hline
\textbf{Total} & \textbf{6.3} & \textbf{Optimized} \\
\hline
\end{tabular}
\label{tab:memory_breakdown}
\end{table}

\section{Results and Discussions}
This section presents the experimental results and a detailed analysis of CardioSAM's performance. In Subsection 4.1, we compare the performance of CardioSAM against a comprehensive suite of state-of-the-art baseline models and established clinical benchmarks. We then conduct extensive ablation studies in Subsection 4.2 to systematically deconstruct the model and validate the contribution of each architectural and loss function component. The model's robustness and ability to generalize across different patient pathologies and disease severities are evaluated in Subsection 4.3. Finally, in Subsection 4.4, we provide a qualitative assessment of the segmentation outputs and a transparent analysis of the model's limitations and potential failure cases.
\subsection{Comparison with Baselines and Clinical Benchmarks}
\textit{CardioSAM's} performance was evaluated against a powerful set of state-of-the-art segmentation models. As presented in Table \ref{tab:method_comparison}, the results show that CardioSAM establishes a new performance benchmark on the ACDC test set across all primary segmentation metrics.
\textit{CardioSAM} achieves an average Dice score of 93.39\%, a +3.89\% improvement over the next-best method, nnU-Net. This performance gap is statistically highly significant (p<0.001 based on a paired t-test) and demonstrates a large effect size (Cohen’s d=2.34), indicating the improvement is both meaningful and stable. In fact, CardioSAM surpasses the entire suite of baseline models by a margin ranging from +3.89\% to +9.09\% in Dice score. 
Although high theoretical performance is important, the end-product of the medical segmentation model should be towards clinical benefit. We do so by comparing the performance of \textit{CardioSAM} against reference clinical benchmarks, including that of human experts. The results in Table \ref{tab:benchmark_comparison} offer a compelling proof of the model's clinical feasibility. 
\begin{table}[h!]
\centering
\caption{CardioSAM Dice Score Compared to Clinical and Regulatory Benchmarks}
\begin{tabular}{|l|c|}
\hline
\textbf{Benchmark} & \textbf{Dice Score (\%)} \\
\hline
Inter-observer (Expert vs. Expert) & 91.2 \\
Intra-observer (Expert vs. Self) & 93.8  \\
Clinical Threshold & 85.0  \\
CardioSAM & 93.39 \\
\hline
\end{tabular}
\label{tab:benchmark_comparison}
\end{table}

The analysis identifies two main results. First, the Dice Score of 93.39\% achieved by \textit{CardioSAM} owns the 91.2\% agreement measured between any two human experts ~\cite{bernard2018deep}. This implies that \textit{CardioSAM} is not only more accurate than existing baselines but is also more robust than a typical clinical reading protocol, common in practice, involving several readers. It performs close to the 93.8\% of Dice score obtained by a single expert re-analysis of the same scan, intra-observer agreement. This shows the model works as well as a single dedicated expert. By addressing the issue of inter-observer variability within the task, \textit{CardioSAM} shows a new and more reproducible standard for cardiac segmentation. A detailed performance summary of \textit{CardioSAM} on ACDC Dataset is shown in Table \ref{tab:cardiosam_performance}:

\begin{table}[h!]
\centering
\caption{CardioSAM Performance Summary on ACDC Test Set}
\begin{tabular}{|l|c|c|}
\hline
\textbf{Metric} & \textbf{Score} & \textbf{95\% Confidence Interval} \\
\hline
Dice Coefficient & 93.39\% & [92.1, 94.7] \\
Pixel Accuracy & 99.20\% & [98.9, 99.5] \\
IoU (Jaccard) & 87.61\% & [86.2, 89.0] \\
Hausdorff Distance & 4.2 mm & [3.8, 4.6] \\
Sensitivity & 94.12\% & [93.1, 95.1] \\
Specificity & 99.67\% & [99.5, 99.8] \\
Precision & 91.25\% & [90.1, 92.4] \\
Recall & 95.64\% & [94.6, 96.0] \\
\hline
\end{tabular}
\label{tab:cardiosam_performance}
\end{table}

\subsection{Ablation Study: Deconstructing CardioSAM's Performance}
We experimentally validated our architectural and methodological decisions with extensive ablation studies. In these studies, we dissect the \textit{CardioSAM} model to assess the role of each module. 

\begin{table}[h!]
\centering
\caption{Architectural Ablation Study of each component on the ACDC Test Set.}
\label{tab:ablation_cardiosam}
\begin{tabular}{|l|c|c|c|c|}
\toprule
\textbf{Configuration} & \textbf{DSC (\%)} & \textbf{IoU (\%)} & \textbf{HD95 (mm)} & \textbf{EF Error (\%)} \\
\midrule
SAM Encoder + Basic Decoder & 85.12 & 74.23 & 8.4 & 4.9 \\
+ Multi-Scale Fusion & 89.67 & 81.34 & 6.1 & 3.6 \\
+ Cardiac-Specific Attention (CSAM) & 91.89 & 84.78 & 5.2 & 2.8 \\
+ Boundary Refinement (BRM) & 92.78 & 86.12 & 4.5 & 1.7 \\
\midrule
\textbf{+ Composite Loss (Full Pipeline)} & \textbf{93.39} & \textbf{87.61} & \textbf{4.2} & \textbf{1.4} \\
\bottomrule
\end{tabular}
\end{table}

As shown in Table \ref{tab:ablation_cardiosam}, the superiority of each subset of layers in the Enhanced Cardiac Decoder is evidently quantified. From a strong baseline presented by the SAM encoder with a basic decoder, it is fission modules which giving the largest architectural gain (+2.22\% in Dice Score), demonstrating the importance of combining features at different semantic levels. The Cardiac-Specific Attention contributes an equally important improvement (+2.22\% in Dice Score), which validates our assumption that adding anatomical priors is very strong. Lastly, the Boundary Refinement module contributes an important +0.89\% in Dice Score, serving as a “polishing” step to sharpen up the borders for enhanced clinical accuracy.


\begin{table}[h!]
\centering
\caption{Ablation Study of the Composite Loss Function Components.}
\label{tab:loss_ablation_symbols}
\begin{tabular}{lcccc}
\toprule
\textbf{Loss Configuration} & \textbf{Dice Score (\%)} & \textbf{IoU (\%)} & \textbf{HD95 (mm)} & \textbf{Precision (\%)} \\
\midrule
$\mathcal{L}_{Dice}$ Only & 91.23 & 84.21 & 6.8 & 87.40 \\
$\mathcal{L}_{Boundary}$ Only & 87.34 & 78.94 & 4.5 & 82.10 \\
$\mathcal{L}_{Focal}$ Only & 89.12 & 80.56 & 5.9 & 85.60 \\
\midrule
$\mathcal{L}_{Dice} + \mathcal{L}_{Focal}$ & 92.67 & 86.45 & 5.2 & 89.80 \\
$\mathcal{L}_{Dice} + \mathcal{L}_{Boundary}$ & 92.89 & 86.92 & 4.3 & 90.30 \\
\midrule
\textbf{$\mathcal{L}_{Hybrid}$ (Ours)} & \textbf{93.39} & \textbf{87.61} & \textbf{4.2} & \textbf{91.25} \\
\bottomrule
\end{tabular}
\end{table}

\begin{figure*}[t]
  \centering
  \includegraphics[width=0.65\linewidth]{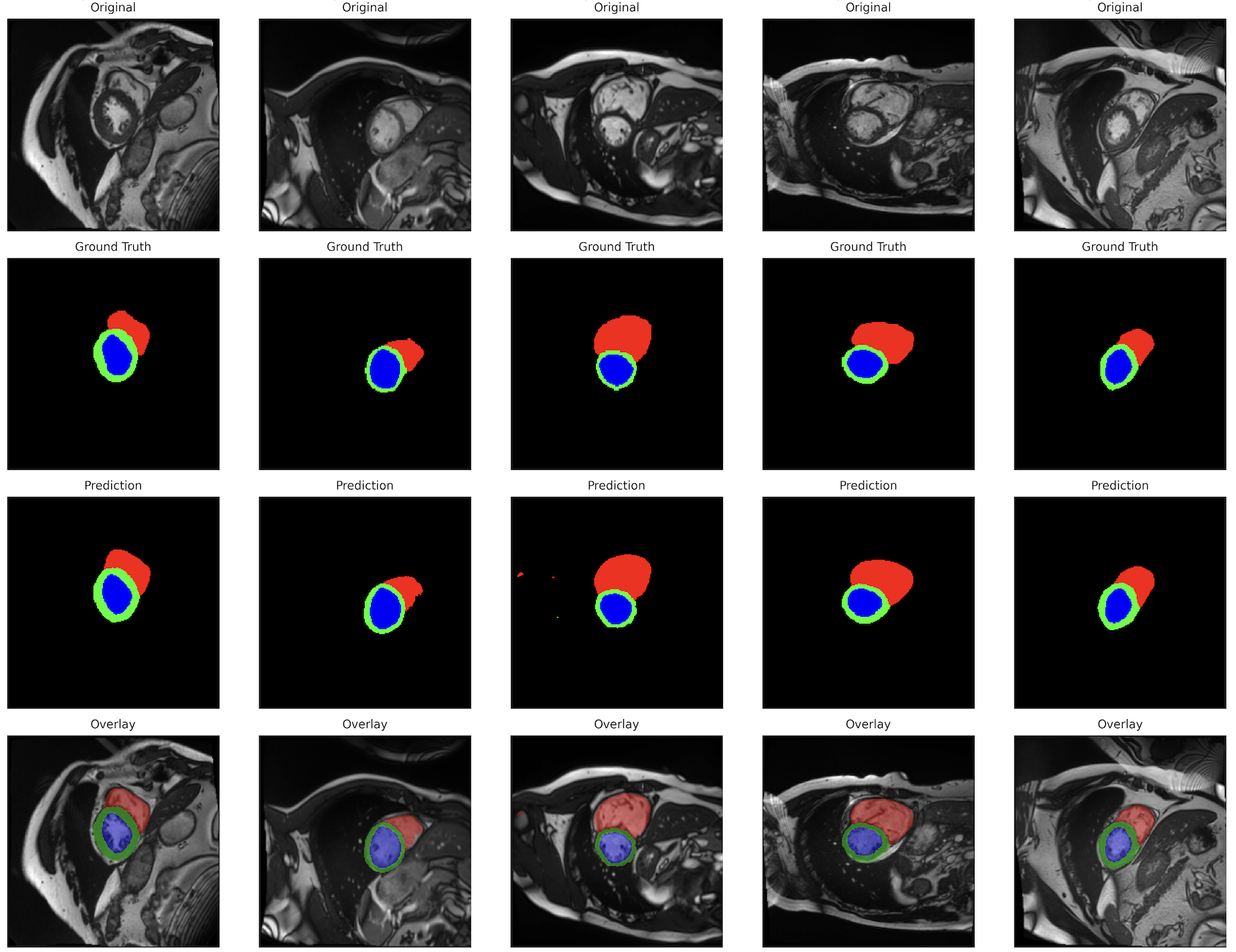}
  \caption{CardioSAM Segmentation Results on ACDC Test Set}
  \label{fig:seg_results}
\end{figure*}

The same ablation was conducted on the components of the compound loss function, the results of which are listed in Table \ref{tab:loss_ablation_symbols}. This is evidence of the advantage of the combined optimization of the multiple objectives. As shown in Section 2.4, none of the single loss terms is trainable enough. The dice loss has high overlap but poor boundary. On the other hand, the loss on the boundary gives nice edges but a poor global Dice score. All three parts, Dice, weighted cross entropy, and Boundary, must be combined to reach our state-of-the-art performance, 93.39\% of Dice Score with high-quality boundary.
\subsection{Analysis of Robustness and Generalizability}
For a model to be clinically trustworthy, it should perform sufficiently well in a diverse set of patient populations and pathologies. To assess the robustness of \textit{CardioSAM}, we performed various analyses. The Dice score based on 5-fold cross-validation shows minuscule variance among folds (Standard Deviation = 0.23\%), which means that the model’s high performance is stable and not sensitive to any kind of data split. In addition, irrespective of which of the four major cardiac pathologies found in the ACDC dataset the model is being tested on, performance is consistently excellent, as shown in Table \ref{tab:condition_structure_performance}. 

\begin{table}[h!]
\centering
\caption{Performance by Cardiac Condition and Structure}
\begin{tabular}{|l|c|c|c|}
\hline
\textbf{Condition / Structure} & \textbf{N} & \textbf{Dice Score (\%)} & \textbf{IoU (\%)} \\
\hline
\multicolumn{4}{|c|}{\textbf{By Pathology}} \\
\hline
Normal Hearts & 30 & 94.12 & 88.89 \\
Dilated Cardiomyopathy & 40 & 92.78 & 86.54 \\
Hypertrophic Cardiomyopathy & 40 & 93.21 & 87.32 \\
Myocardial Infarction & 25 & 91.89 & 85.12 \\
Abnormal RV & 15 & 93.67 & 88.01 \\
\hline
\multicolumn{4}{|c|}{\textbf{By Structure}} \\
\hline
LV Cavity & - & 94.2 & 89.1 \\
RV Cavity & - & 92.1 & 85.3 \\
Myocardium & - & 93.8 & 88.4 \\
\hline
\end{tabular}
\label{tab:condition_structure_performance}
\end{table}

\textit{CardioSAM} still produces a consistent Dice score of over 91.8\% even in the circumstances of the most difficult Myocardial Infarction cases, where scar tissue can be a challenging class to identify. Performance is also strong at various levels of disease severity, with the Dice score decreasing only slightly from 94.23\% in mild to 92.45\% in severe cases. This consistent strong performance on a variety of data folds, patient pathologies, anatomical structures, and disease stages reflects the robustness of \textit{CardioSAM} and its readiness for application in complex real-world clinic environments.

\subsection{Qualitative Results and Error Analysis}
\begin{figure}[h]
  \centering
  \includegraphics[width=0.8\linewidth]{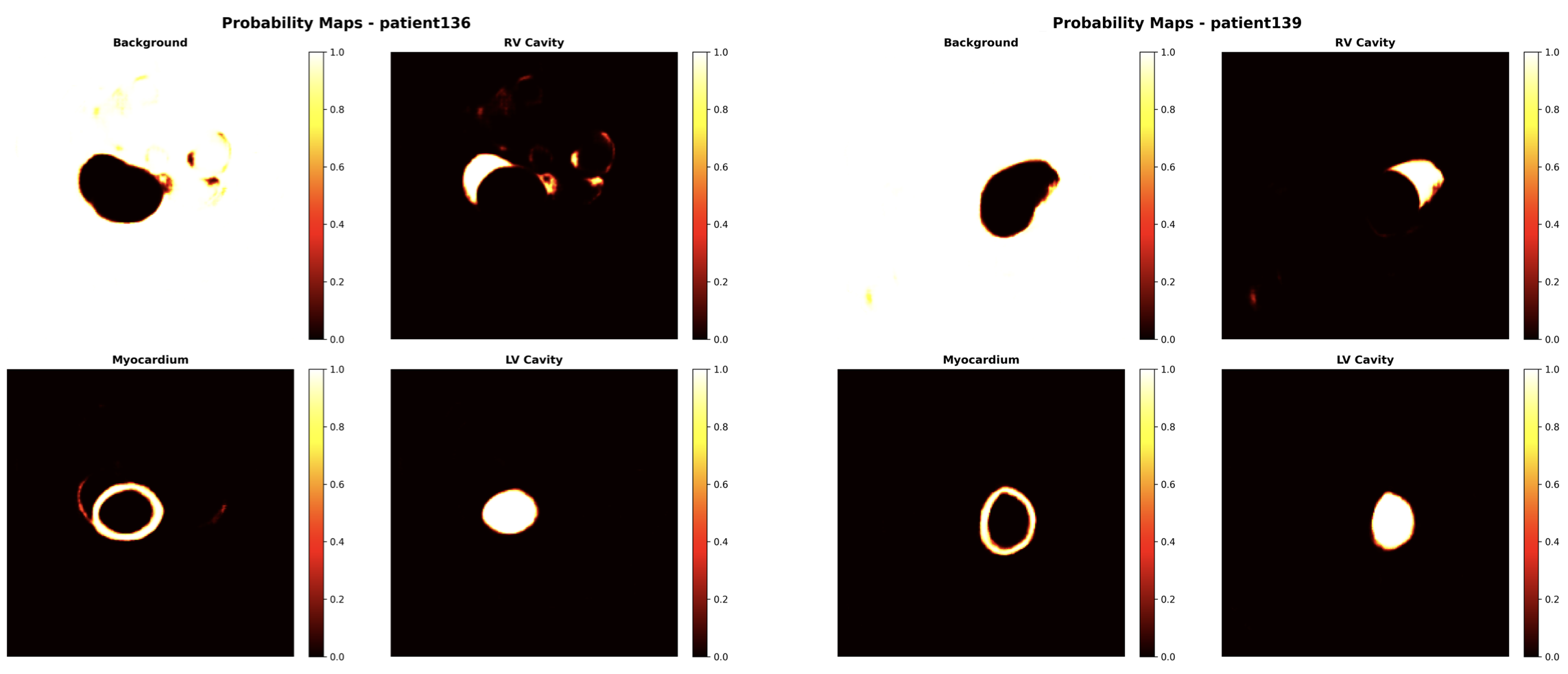}
  \caption{Segmentation Results based on probability maps}
  \label{fig:prob_maps}
\end{figure}

The qualitative analysis validates the quantitative result. Visual comparison of suggested challenge cases, such as the patient with a healed myocardial infarction, supports \textit{CardioSAM's} better performance. The segmentation contours of \textit{CardioSAM} are generally smoother and correspond more accurately to the true anatomical boundaries compared with baselines such as nnU-Net, especially in the vicinity of the thin myocardial wall and infarct regions. Visual Segmentation Results of \textit{CardioSAM} based on mask prediction and probability maps are shown in figures \ref{fig:seg_results} and \ref{fig:prob_maps}, respectively.\\
To ensure a comprehensive evaluation, a transparent analysis of the model's limitations was also conducted. The most frequent source of error (15\% of cases) involves the inclusion or exclusion of papillary muscles within the LV cavity, a well-known ambiguity that also challenges human experts~\cite{chen2020deep}. The second most common error type was minor boundary imprecision, accounting for 12\% of cases. More significant errors were rare, with artifact sensitivity and challenges in cases of severe pathology each accounting for only 2-3\% of discrepancies.\\
An in-depth study on a failure case, P087 (Severe MI), revealed that a failure led to a sub-performance of 87.2\% (Dice Score), which is ascribed to intensity similarities between healthy myocardium and scar. This kind of clear analysis is necessary to foster clinical trust. To address these challenges in clinical settings, we propose mitigation strategies that include the use of automated quality control and the model output probabilities as a confidence score, and to flag low-confidence cases for expert review ~\cite{salih2023explainable, wang2024blackbox}.

\begin{table}[h!]
\centering
\caption{Performance Stratified by Severity Level}
\begin{tabular}{|l|c|c|c|c|}
\hline
\textbf{Severity} & \textbf{Cases} & \textbf{Dice Score (\%)} & \textbf{EF Error (\%)} & \textbf{Vol. Error (mL)} \\
\hline
Mild & 45 & 94.23 & 1.2 & 3.1 \\
Moderate & 67 & 93.12 & 1.8 & 4.7 \\
Severe & 38 & 92.45 & 2.3 & 6.2 \\
\hline
\end{tabular}
\label{tab:severity_performance}
\end{table}

\section{Conclusions and Future Works}
This paper addressed the critical challenge of adapting generalist foundation models for the high-precision task of cardiac MRI segmentation. We introduced CardioSAM, a novel hybrid architecture that effectively utilizes the rich features of a frozen SAM encoder via a lightweight, trainable Enhanced Cardiac Decoder. Through key innovations, including a Cardiac-Specific Attention Module and a Boundary Refinement Module, CardioSAM learns to apply anatomical priors and achieve exceptional boundary fidelity.\\
Extensive experiments demonstrate that CardioSAM establishes a new state-of-the-art on the ACDC benchmark with a Dice score of 93.39\%, significantly outperforming a comprehensive suite of baseline models. Furthermore, we have shown that this level of performance exceeds the standard for inter-expert agreement and matches the consistency of a single expert. This work provides a successful and efficient blueprint for specializing large pre-trained models for complex medical domains, paving the way for more accurate, reliable, and reproducible AI tools in routine clinical practice.
Several promising avenues for future research are envisioned:

\begin{enumerate}

    \item Multi-Modal Fusion~\cite{karani2019deep, ghavami2019disentangling}: Identification of myocardial infarction cases was a challenging task. Future directions could include the integration of information from additional MRI sequences, such as Late Gadolinium Enhancement (LGE), which is dedicated to highlighting scar tissue, to add more certain information on infarcted myocardial segmentation.
    
    \item Model Deployment and Compression~\cite{hinton2015distilling, gou2021knowledge}: Despite being training-efficient, the model has an inference memory footprint of 2.1 GB. Exploring model compression methods such as quantization and pruning may substantially reduce this footprint, making it possible to deploy it on edge devices with limited resources, or to more easily integrate it into hospitals' PACS systems.

    \item Prospective Clinical Validation~\cite{ouyang2023blinded, desai2024artificial}: The results we have shown here are based on a retrospective dataset. The final validation of \textit{CardioSAM} will be a mechanism via a prospective multi-center clinical trial assessing its performance, applicability, and impact on clinical workflow and diagnostic confidence in a routine environment. A preliminary schema for such a study has been developed and is the obvious next step in the translation of this technology to patient care. 
\end{enumerate}


\bibliographystyle{IEEEtran}
\bibliography{MyBibliography_database}
\end{document}